# Watson & Holmes: A Naturalistic Benchmark for Comparing Human and LLM Reasoning

Thatchawin Leelawat[1] & Lewis D Griffin[2]

Computer Science, University College London, UK

[1]`thatchawin.leelawat.19@ucl.ac.uk` [2]`L.Griffin@cs.ucl.ac.uk`

Existing benchmarks for AI reasoning provide limited insight into how closely these capabilities resemble human reasoning in naturalistic contexts. We present an adaptation of the *Watson & Holmes* detective tabletop game as a new benchmark designed to evaluate reasoning performance using incrementally presented narrative evidence, open-ended questions and unconstrained language responses. An automated grading system was developed and validated against human assessors to enable scalable and replicable performance evaluation. Results show a clear improvement in AI model performance over time. Over nine months of 2025, model performance rose from the lower quartile of the human comparison group to approximately the top 5%. Around half of this improvement reflects steady advancement across successive model releases, while the remainder corresponds to a marked step change associated with reasoning-oriented model architectures. Systematic differences in the performance of AI models compared to humans, dependent on features of the specific detection puzzle, were mostly absent with the exception of a fall in performance for models when solving longer cases (case lengths being in the range of 1900-4000 words), and an advantage at inductive reasoning for reasoning models at early stages of case solving when evidence was scant.

# 1. Introduction

Reasoning has long been regarded as a defining feature of human intelligence; and, more recently, artificial intelligence (AI) [1]. From early symbolic and rule-based systems to today's large language models (LLMs), progress in AI has often been measured by the extent to which machines can replicate or approximate human reasoning [2].

Unlike previous engineered reasoning systems, which follow explicit logical rules, contemporary AI models acquire reasoning capabilities implicitly through exposure to massive text corpora ; emerging [2] as a by-product of language prediction [3]. Exploring how such reasoning arises and how it can be reliably elicited or improved has become essential for understanding the boundaries of LLM intelligence [4].

Several benchmarks have been developed to track the rapid advancement of AI reasoning capabilities across diverse domains. In the reasoning field, these benchmarks assess a wide range of sub-types, including Commonsense, Mathematical, Logical and Deductive, Algorithmic and Symbolic, Domain-specific and Multi-hop. We have adapted the 'Watson & Holmes' (W&H) tabletop game [5] to create the WnH benchmark for AI reasoning. W&H is a whodunnit type game, in which players compete to be the first to solve a mystery using reasoning skills such as deduction, induction, and abduction.

The aims of the WnH benchmark are the following:

    1. To measure AI reasoning performance on naturalistic tasks expressed in rich but ordinary language, requiring wide but non-specialist knowledge and a range of reasoning modes.

    2. To measure AI reasoning using an unconstrained answer format that encourages and tests open-ended reasoning.

    3. To allow comparison of a range of AI models including both completion models (cModels) and reasoning models (rModels).

    4. To track the progress of AI reasoning capabilities over time and relative to compute cost.

The benchmark will support answering the research questions:

    1. How does AI reasoning in a naturalistic, open-ended task differ from human reasoning performance?

    2. How does AI performance on the Watson & Holmes benchmark compare to performance on other established reasoning benchmarks?

    3. Do features of specific reasoning tasks influence AI performance compared to humans?

# 2. Background

This section outlines key perspectives and developments in reasoning and its evaluation. It reviews major theoretical accounts of human and machine reasoning, surveys existing reasoning benchmarks used to assess AI models' reasoning capabilities across domains, examines approaches to autograding open-ended reasoning performance, and concludes with a discussion of current benchmark limitations and gaps.

## 2.1. Background on Reasoning

Human reasoning has been a central focus of inquiry across philosophy, psychology, mathematics, and artificial intelligence. In these disciplines and in specific domains, researchers have proposed diverse theories to understand how people draw conclusions, solve problems, and make decisions.

### 2.1.1. Philosophical perspectives

Philosophical approaches to reasoning typically present idealized accounts of how reasoning should operate to draw conclusions. A standard account is the classification of reasoning modes into deduction, induction and abduction. Deduction and induction were already recognized and distinguished in ancient Greek philosophy, particularly in Aristotle's logic [6]. Much later in the 19th century Charles Sanders Peirce [7] added abduction to this framework. The modes can defined as follows:

- Deduction: Reasoning following established canonical logical patterns.
- Induction: Inferences based on generalization from previous, ideally repeated, observations.
- Abduction: The best possible explanation for the given facts.

### 2.1.2. Psychological perspective

From a psychological standpoint, reasoning is concerned with how people actually think and make decisions. In contrast to philosophical accounts of idealized reasoning, psychological theories emphasize that human reasoning is shaped by cognitive processes, heuristics, mental models, and social dynamics rather than being purely logical.

Psychologists have studied the role of heuristics, mental shortcuts that simplify decision-making. The classic heuristics and biases program [8] hypothesized that human reasoning relies on such heuristics. This hypothesis was motivated by the observation of systematic patterns of judgment errors that were otherwise difficult to explain under models of fully rational reasoning. While heuristics enable robust and efficient decision-making in complex, real-world environments, they also lead to predictable deviations from normative logic and probability [9].

Also influential is the mental model theory [10] which proposes that human reasoning often relies less on application of rules and more on internal simulations of possible situations. Modern formulations integrate probabilistic reasoning and computational constraints, acknowledging that reasoners may not always construct full sets of possible models [11–13].

A prominent contemporary framework is the Dual Process Theory [14], which divides reasoning into type 1 and type 2. Type 1 is quick, instinctive, and often relies on mental shortcuts. It can be influenced by personal biases due to pattern recognition or past experience. Type 2 works at a slower pace, engaging in careful and logical analysis. Type 2 is engaged when we carefully work through a problem step by step, such as solving a mathematical problem, or making a complex decision by weighing pros and cons.

### 2.1.3. Mathematical Reasoning

Reasoning has also been discussed in domains where it is particularly important; especially mathematics, where it is used to discover, analyze, and prove mathematical truths. Polya [15] outlined a practical four-step problem-solving process: comprehend the problem, formulate a plan, implement the plan, and evaluate the solution. These techniques, while not guaranteeing a solution, can "speed up the process of reaching a satisfactory solution" by guiding one's reasoning in fruitful directions. Pólya also catalogued many heuristic strategies to support reasoning including: drawing a diagram to visualize the problem, examining special cases or examples, considering related problems, and breaking the problem into sub-problems.

Pólya's mid 20$^{th}$ century notion of heuristic was explicitly intended as a guide for human minds in classrooms and research, not for mechanical or computational application, but recently, mathematicians have started to explore the limits and opportunities for AI reasoning for mathematics. Gowers [16] has argued that artificial intelligence may eventually perform human-style mathematical reasoning, rather than being limited to pattern-matching or statistical learning. Initiatives such as the Polymath Project (2009) [17] demonstrated how large-scale collaboration, guided by human heuristics, could collectively solve difficult mathematical problems. In parallel, the rise of formal proof assistants and autograders [18, 19] provided systems that verify logical correctness, enabling collaboration and rigor at scale.

### 2.1.4. AI Reasoning

AI reasoning is not designed explicitly to replicate human cognition [20], but the methods used to elicit reasoning from AI models are often cognitively inspired. For example, Chain-of-Thought (CoT) prompting [3], asks a model to "think step by step" before providing an answer [3, 21]. This technique has some alignment with deductive and inductive reasoning, and with type 2 reasoning. CoT significantly improves AI capabilities on various benchmarks such as GSM8K [3]. Building on this idea, researchers proposed lighter variants such as Chain-of-Draft [22], which augments CoT with concise reasoning prompts. Instead of long rationales, the model is asked to generate a short and bounded reasoning step (limit to five words per step). This causes the model to achieve performance close to full CoT while moderating token usage.

CoT performance can be further improved through self-consistency [23], which samples multiple reasoning paths and aggregates their conclusions, thereby reducing spurious errors. Beyond linear chains, Tree-of-Thought prompting [24] generalizes CoT by allowing branching exploration and backtracking, a process that closely resembles human problem-solving through hypothesis testing. Other cognitively inspired approaches include least-to-most prompting [25], where the model decomposes a complex task into simpler subproblems before solving them sequentially, and role-play prompting [26], where models adopt a role to provide scaffolding to their reasoning.

Later, a new generation of reasoning models emerged that incorporate internal mechanisms to support multi-step inference more directly. A notable example is DeepSeek-R1, which was trained using reinforcement learning objectives that reward not only correct final answers but also the production of structured and readable reasoning traces [27]. Such innovations mark a shift from completion-only models (cModels) towards reasoning models (rModels). Similar developments can be observed across the industry: OpenAI's o1 and o3 series reportedly extend inference-time computation to scale internal reasoning depth [28, 29], while Google's Gemini 2.5 Pro/Flash models also emphasize enhanced multi-step reasoning capabilities [30]. These advances illustrate how AI models are beginning to embed structured reasoning processes within their architecture and training regimes, moving beyond reasoning elicited purely through prompting.

## 2.2. Reasoning Benchmarks

Tasks used to evaluate AI capabilities are known as benchmarks and have been developed to assess a wide range of competences. Here we focus on reviewing benchmarks related to reasoning, before reviewing some common implementation issues such as their reliance on multiple-choice formats, tainting of results due to data leakage, and the lack of consistent human baselines for comparison.

Some reasoning benchmarks are constructed around a particular domain of human activity and use a task that tests a mixture of reasoning, domain understanding and domain knowledge, while others aim for varying degrees of abstract reasoning minimizing the role of knowledge. We review the following types: Commonsense Reasoning, Mathematical Reasoning, Logical and Deductive Reasoning, Algorithmic and Symbolic Reasoning, Domain-specific Reasoning, and Multi-hop Reasoning.

### 2.2.1. Commonsense Reasoning

Common Sense Reasoning Benchmarks aim to assess AI reasoning about situations from everyday life. Notable examples include:

**CommonsenseQA** [31] is ~12K multiple choice questions (MCQs) designed to test everyday commonsense knowledge that is obvious to humans but challenging for AI models. For example "Where on a river can you hold a cup upright to catch water on a sunny day?" with answer options: waterfall (correct), bridge, valley, pebble, and mountain.

**WinoGrande** [32] is a 44K MCQ pronoun resolution dataset, evaluating AI models ability to determine which noun a pronoun refers to within a sentence. For example: "The trophy doesn't fit into the brown suitcase because **it** is too large" has the correct answer 'trophy' and incorrect answer 'suitcase'.

**HellaSwag** [33] is an MCQ commonsense reasoning benchmark where the most plausible sentence to complete a given paragraph must be chosen. These paragraphs are taken from video caption datasets featuring daily activities and step-by-step instructional articles from the WikiHow website.

**PIQA** [34] is a dataset of 21K MCQs dataset focused on everyday actions, including practical reasoning and knowledge of objects' shape, material, and intended use. Each question follows a goal–solution format, where the goal describes an intended action and the model must choose the more plausible solution. The dataset draws inspiration from Instructables.com.

**Social IQa** [35] is a dataset of 38K MCQs focused on everyday social scenarios. Question types include understanding a person's wants, reactions, descriptions, motivations, needs, and effects.

### 2.2.2. Mathematical Reasoning

Benchmarks of this type assesses an LLM's ability to understand and solve mathematical problems. The problems can be of different complexity, ranging from elementary-level exercises to advanced, competition-grade challenges.

At the elementary level, common benchmarks include GSM8K [36], which contains 8,500 grade school math problems, and its simplified version, GSM1K [37]. SWAMP [38] offers similar elementary-level problems, but this dataset introduces more variation in sentence structures. Other benchmarks in this category include CMATH [39], which measures the elementary reasoning of LLMs in the Chinese language, and VNHSGE [40], which is based on the Vietnamese High School Graduation Examination.

At the high school level, MATH [41] includes 12,500 American competition problems spanning seven mathematical domains, such as AMC and AIME. Another benchmark, AGIEval SAT-MATH [42], is derived from SAT math problems, and JEEBench [43] presents Indian joint entrance examination questions.

At the higher education level, UGMathBench [44] provides ~5K undergraduate-level questions for evaluation, while U-MATH [45] focuses on open-ended university mathematics problems. For graduate-level reasoning, HARDMATH [46] targets skills in asymptotic thinking and approximation.

### 2.2.3. Logical and Deductive Reasoning

This type of reasoning follows formal rules, draws logically valid inferences, and maintains internal consistency. Natural Language Inference (NLI) is a key area within logical reasoning, where models determine whether a hypothesis is entailed, contradicted, or neutral with respect to a given premise [47–50]. Several benchmarks have been developed to evaluate this ability. For example, ConjNLI [47] challenges models to reason about conjunctions within sentences. TaxiNLI [48] focuses on logical reasoning across categories such as mathematical logic, connectives, and deductive inference. ConTRoL [50] emphasizes contextual reasoning, requiring models to make inferences based on nuanced contextual cues. Additionally, HELP [49] and MED [51] are designed to assess monotonicity, a core principle in formal logic that tests how conclusions are preserved under specific operations.

Other benchmarks that fall into this category include ReClor [52] and LogiQA [53], both of which are reading comprehension datasets that require logical reasoning. LogiQA features MCQs derived from Chinese civil service and law school exams, translated into English, and covers logical puzzles and analytical reasoning tasks. ReClor, on the other hand, is constructed from logical reasoning questions in standardized exams (such as the LSAT). It presents passages followed by questions that involve identifying assumptions, recognizing logical fallacies, or completing arguments.

### 2.2.4. Algorithmic and Symbolic Reasoning

This category focuses on working with symbols and following clear step-by-step rules, especially in ways that are not purely numerical. It emphasizes logical thinking, solving problems like a computer, and handling patterns or structures in language or code.

Benchmarks in this category range from simpler code generation and problem-solving tasks such as HumanEval [54], which focuses on logic, syntax, and correctness, to more advanced challenges such as MBPP [55] designed to test procedural reasoning and generate correct code using introductory programming tasks. In the advanced class, APPS [56] targets broad range of programming problems difficulties, from basic to competitive-level. CodeContests [57] focus specifically on competition-grade programming problems sourced from real contests like Codeforces, demanding deep logical thinking and algorithmic design. An extension, CodeContests+ [58], generates additional AI-curated problems and test cases. QuixBugs [59], focuses on automatic bug correction in code.

### 2.2.5. Domain-specific Reasoning

Domain-specific reasoning benchmarks test models on questions that require knowledge related to a particular domain: scientific, legal or medical. This assesses both reasoning, and knowledge retrieval and application.

Humanity's Last Exam (HLE) [60] is a 2.5K question multimodal benchmark that includes both text-only items and questions paired with reference images. HLE positions itself as testing performance at the frontier of expert human knowledge. It spans mathematics, humanities, natural sciences, and computer science. HLE consists of multiple-choice and short-answer questions graded using exact-match evaluation. Questions were contributed by experts and rigorously filtered and reviewed including automated difficulty checks against frontier language models and subsequent human expert review. These questions are intentionally crafted to be difficult, precise, and resistant to simple internet search or retrieval. According to the public HLE leaderboard, the current top model achieves approximately 38% accuracy [61, 62].

**Science-Based Benchmarks**

The AI2 Reasoning Challenge (ARC) [63] consists of ~8K MCQs on science topics from school grades 3-9. Models are required to recall factual knowledge and apply reasoning based on those facts. This benchmark focuses on recall, reasoning, knowledge synthesis, and inference.

OpenBookQA [64] also focuses on the science domain, but requires models to extract relevant information from a provided collection of scientific facts rather than retrieve from memory. To solve these questions, models must combine those extracted facts with general reasoning skills.

SciQ [65] contains ~14K science MCQs gathered through crowdsourcing. Unlike ARC and OpenBookQA, SciQ emphasizes the recall of factual knowledge and basic comprehension of science topics rather than complex reasoning or inference.

**Biomedical and Medical Benchmarks**

In more specialized scientific domains, benchmarks such as PubMedQA [66] and MedQA [67] have been developed. PubMedQA evaluates models on biomedical question answering, requiring understanding of scientific research texts in the form of PubMed abstracts. MedQA, in contrast, focuses on the medical domain and includes questions based on medical licensing exams. These benchmarks demand a high degree of domain knowledge and the ability to interpret and reason over complex scientific and clinical information.

Recently, the MedXpertQA [68] benchmark has been proposed for expert-level medical reasoning. It includes ~4K questions and introduces a multimodal variant that combines text with clinical images. The dataset is adapted from real patient records and examination results, making it more clinically realistic than earlier benchmarks.

Another advancement is LLMEval-Med [69], which emphasizes evaluation on real-world clinical data. It contains ~3K questions derived from electronic health records, spanning five essential medical dimensions with a focus on open-ended tasks rather than only multiple choice. LLMEval-Med also introduces a dynamic evaluation framework that combines automated "LLM-as-Judge" assessment with expert-developed checklists, iteratively refined through human–machine agreement analysis to ensure reliability. Similarly, MedQA-CS [70] adopts an AI-Structured Clinical Examination approach, modeled after OSCEs, where the LLM functions as an automated examiner to assess professional, open-ended clinical responses.

**Legal Reasoning Benchmarks**

LegalBench [71] is an open-source benchmark developed collaboratively by experts in law and computer science. It contains 162 diverse tasks aimed at evaluating AI capability on reasoning in legal contexts. The tasks are of six types: issue-spotting (identifying whether a scenario raises a legal issue), rule-recall (recognizing the correct legal rule), rule-application (applying a legal rule to a set of facts), rule-conclusion (deriving the legal outcome), interpretation (extracting meaning from legal texts), and rhetorical-understanding (figuring out the role or purpose of language in legal arguments). From the perspective of classical reasoning, the legal domain relies most heavily on deduction, particularly in rule-application and rule-conclusion, and on abduction, which is central to issue-spotting, interpretation, and rhetorical-understanding.

### 2.2.6. Multi-hop Reasoning

Multi-hop reasoning benchmarks assess how well a model can connect multiple pieces of information to deduce a conclusion. These tasks often require retrieving or inferring intermediate steps (hops) before concluding.

One commonly used benchmark is HotpotQA [72], a large-scale crowdsourced dataset containing 113K questions specifically designed for multi-hop question answering. Solving these questions requires identifying relevant information from the provided paragraphs and synthesizing multiple facts to conclude.

Unlike other multi-hop reasoning benchmarks, StrategyQA [73] is a dataset of ~3K multi-step questions that emphasizes implicit reasoning, where the reasoning step is not stated in the question or the data. It challenges models to infer the necessary intermediate steps on their own to answer yes/no questions.

MuSiQue [74] contains 25K questions requiring 2-4 reasoning hops. These questions are carefully constructed by combining single-hop questions—a method designed to reduce reasoning shortcuts. This approach helps mitigate issues such as overly specific sub-questions or train-test leakage.

SCoRE [75] extends multi-hop reasoning into the domain of commonsense knowledge, covering the spatial, temporal, social, and natural domains. Rather than retrieving textual evidence, SCoRE evaluates whether models can perform structured, multi-step reasoning by composing multiple commonsense statements to reach a logical conclusion.

### 2.2.7. Abstract Reasoning

ARC-AGI-1 [76] is a visual benchmark designed to test the ability of systems to rapidly acquire unfamiliar skills and solve problems that are not explicitly included in their training data. Each task is presented as small (e.g. 20 by 20) grids of coloured cells organized as a few pairs of input–output demonstrations. The system must infer the underlying transformation rules from input to output from these demonstrations and apply them to novel inputs. Because the rules are never provided directly, solutions cannot be achieved through memorization, and the benchmark is designed to avoid leakage of test data into model training. While human participants generally find the tasks solvable (98% success), even advanced machine learning systems continue to struggle with them (o3-preview (low) model achieves only 76% accuracy) [77].

Later, ARC-AGI-2 [78] was introduced with several refinements: reducing susceptibility to brute-force solutions, incorporating large-scale human testing to establish reliable baselines, expanding the task set to cover a wider range of difficulty, and calibrating challenge levels across the public, private, and semi-private evaluation splits to ensure consistent human solvability.

A key strength of ARC-AGI benchmarks is that they do not require domain-specific or world knowledge, and are resistant to data leakage. However, their narrow focus on abstract tasks raise questions about the extent to which performance on these benchmarks can generalize to broader reasoning tasks.

### 2.2.8. Game Based Benchmarks

A growing trend in LLM benchmarking is adapting and modifying games to serve as reasoning benchmarks. This approach is promising, as the structured rules and clear objectives of games provide a controlled environment to evaluate reasoning, strategy, and decision-making capabilities. Additionally, compared to static benchmarks, interactive game-based evaluations more closely resemble real-world reasoning, where models must adapt, plan, and respond dynamically during the interaction. Notable game-based benchmarks are the following:

GTBench [79] includes classic games such as Tic-Tac-Toe, Connect Four, Negotiation, and Nim, used to evaluate the mathematical, algorithmic, and logical reasoning of AI models; and modern competitive games like Diplomacy, Werewolf, and Avalon, which assess capabilities such as deception detection, role inference, coalition formation, and negotiation. Both classes of game require multi-hop, interactive, and strategic reasoning.

GameArena [80] is a benchmark adapted from three interactive games: Akinator, Taboo, and Bluffing. Each game is designed to evaluate a specific type of reasoning in LLMs.

Akinator is a version of the classic 20-questions game in which an object must be identified by asking a series of questions with answer yes, no or uncertain. This game has been claimed to assess deductive reasoning but may be better characterized as abduction-dominated, since the model must infer the most plausible object from incomplete and often uncertain responses, updating its hypotheses as new information becomes available.

In the Taboo game, the goal of the human player is to ask the AI model questions that lead it to say a chosen 'taboo' word, while the model attempts to guess that word before it utters it in a response to a question. This game is claimed to engage abductive reasoning, as the model must infer the most plausible concept from indirect and incomplete linguistic cues.

In the Bluffing game, the human participant begins by making a statement that may be accurate or fabricated and attempts to convince the AI model of its credibility. This task engages inductive reasoning, since the model generalizes from a limited set of conversational cues to assess the likelihood of honesty or deception, but also abductive reasoning, as the model infers the most plausible explanation for inconsistencies or behavioral patterns in the player's answers.

While GameArena provides a structured framework for assessing different forms of reasoning in LLMs, it also presents several challenges. The reasoning categories assigned to each game (deductive, inductive, and abductive) are not always clearly delineated, as many tasks involve overlapping cognitive processes.

### 2.2.9. Detective Game based benchmarks

Detective-style benchmarks require models to analyse narrative contexts, identify relevant clues, and infer the most plausible conclusion. The tasks combine elements of logical, multi-hop, and commonsense reasoning, offering a closer simulation of human-style problem solving under uncertainty in naturalistic scenarios.

DetectBench [81] is a benchmark consisting of ~4K MCQs devised around short detective-style scenarios. The benchmark is organized into three levels of difficulty. The baseline DetectBench task uses contexts averaging 140 words, while DetectBench-Hard and DetectBench-Distract introduce longer contexts. Each participant answered 15 questions drawn from a subset of 250 development-set questions, ensuring that each question was independently answered by three participants. Humans achieved 74% accuracy on questions, while the best LLM accuracy achieved was 62% achieved by GPT4 guided by a crafted prompt.

True Detective [82] is a MCQ benchmark in which each case is presented as a detective-style puzzle of around 1,200 words, requiring the culprit to be identified. The benchmark question data are sourced from an online puzzle platform, raising the possibility that some instances may have been exposed during LLM training. Human performance is inferred from solve-rate statistics reported on the same platform; however, the number of participants and their demographic or educational backgrounds are not specified. Solve rate is used as the primary evaluation metric. Human accuracy ranges from 40% to 60%, while GPT-4 achieves 38% under standard conditions.

WHODUNIT [83] is a new benchmark for deductive reasoning in narrative contexts, requiring an AI to identify the culprit in mystery-style stories (median length ≈ 220 pages) directly drawn from published fictions from famous writers such as Agatha Christie and Arthur Conan Doyle. The dataset also includes name-swap variants aimed at defeating memorization. After reading the narrative, the model is prompted with an open-ended question such as 'Who is the culprit?', and its answer is evaluated against the annotated ground truth. Unlike DetectBench and True Detective, which use MCQ, this benchmark evaluates the LLM reasoning by checking whether the model's output exactly matches the culprit's name. GPT-4o correctly identified the culprit in 84% of stories, with performance decreasing by approximately 1–3% with substituted names. Comparison human performance data has not been provided by the study authors.

## 2.3. Autograding of performance

Many benchmarks use MCQs but these have limitations that reduce their usefulness as a gauge of reasoning ability. Their constrained-choice format limits opportunities for open-ended reasoning, synthesis, and justification [84], which are considered important in educational measurement research [85, 86]. Other studies [86, 87] find that when models shift from MCQs to open-ended reasoning tasks, their performance sharply declines. This further highlights the limitations of MCQ-based benchmarks in accurately measuring reasoning ability.

Furthermore, MCQs enforce binary scoring preventing awarding partial credit for partially correct reasoning answers. As observed in commonsense reasoning benchmarks, some MCQs embed ambiguity or multiple plausible answers, meaning that forcing a single "correct" choice can misrepresent nuanced judgments or reasoning paths [88]. Given these limitations of the MCQs, open-ended questions provide a more faithful measure of reasoning ability by allowing systems to articulate their own reasoning processes without being affected by options in MCQ.

While open-ended questions are better probes, their grading is time-consuming and potentially variable. Grading rubrics are commonly used in educational measurement to reduce these issues. Rubrics provide structured evaluation criteria, often encompassing dimensions such as factual correctness, logical coherence, justification quality, and completeness of reasoning [89, 90]. This structured approach enables evaluators to move beyond binary correctness and capture richer aspects of performance, such as whether a response demonstrates sound reasoning, even if the final conclusion is partially incorrect.

Reliable evaluation of open-ended reasoning tasks requires scalable, consistent autograding, since human grading introduces bias, order effects, limited reproducibility across studies, and does not scale to modern benchmark sizes [19, 91]

Early autograding approaches primarily relied on similarity-based methods such as semantic overlap, n-gram matching or embedding similarity to compare responses with a reference answer [87, 92–94]. While effective for factual questions, these techniques fail to capture reasoning quality or partial correctness when responses deviate lexically but remain semantically valid.

Recent work explores open-ended autograding using AIs as automated evaluators. AI-as-judge frameworks assess responses in context, interpreting arguments, explanations, and evidence expressed in free-form text [69, 86]. With well-designed prompts or rubrics, these systems achieve accuracy comparable to human graders while maintaining scalability and reproducibility across large datasets [90, 91, 95]. Such approaches help reduce variability among evaluators, enhance scalability, and promote consistency.

## 2.4. Discussion of benchmarks

Despite their diversity, many reasoning benchmarks share the following limitations. First, most rely on MCQ formats, which simplify evaluation but fail to capture open-ended reasoning and explanation. This restricts assessment to answer accuracy rather than the quality of the reasoning process. For instance, several commonsense reasoning benchmarks such as CommonsenseQA [31], SocialIQA [35], and PIQA [34] employ MCQs to test different aspects of commonsense reasoning. While these datasets are effective for measuring factual correctness, they often overlook partially correct or nuanced reasoning that large language models (LLMs) may produce. DetectBench [81] exemplifies this limitation: although it aims to evaluate evidence detection and reasoning, it still depends on MCQs, constraining its ability to assess explanatory reasoning.

Second, some benchmarks may be prone to leakage of test data into training data, since their questions are sourced from widely available online platforms or classic texts [82, 83] that may appear in pretraining corpora. Leakage would cause reasoning performance to be over-estimated. For instance, in the WHODUNIT benchmark GPT-4o correctly identified the culprit in 84% of stories, with performance decreasing by approximately 1–3% when character names were substituted or the story theme was modified. This reduction suggests that the model had been exposed to the original material during training, and there is no guarantee that simple name substitution is sufficient to eliminate leakage effects fully.

Third, provision of human baselines for comparison is inconsistent. For example, MedQA reports model performance relative to the average passing performance of human medical students on the USMLE; while MedQA-CS positions AIs as "virtual students" in OSCE-style tasks but does not provide direct human student comparison. When human comparisons are provided it is important that the testing pool is well characterized and appropriately motivated.

A further crucial issue is the persistent gap between task competence and real-world competence. Performance on narrowly defined benchmark tasks does not necessarily generalize to reasoning ability in naturalistic settings. Increasing the naturalness and diversity of benchmarks, particularly through interactive or game-based formats, offers one way to address this limitation. While seeming less serious than traditional test-style evaluations, such formats provide a more naturalistic context for assessing reasoning. For instance, GTBench [79] and GameArena [80] evaluate reasoning in interactive, dialogue-driven, or strategic contexts, while detective-style benchmarks such as DetectBench [81] and WHODUNIT [83] simulate narrative problem-solving that more closely reflects real-world inference and decision-making.

Another critical issue with current benchmarks is that many established reasoning datasets, such as CommonsenseQA [31], WinoGrande [32], HellaSwag [33], and PIQA [34], have become saturated with model scores reaching or exceeding human scores. Saturated benchmarks are still useful, for example to assess the performance of lower compute models, but they become imprecise as scores reach the maximum possible.

The ideal reasoning benchmark would combine several key attributes:

- it would be immune to test data leakage,
- provide a precise measure of performance
- be calibrated against robust human baselines,
- resist rapid saturation,
- reflect reasoning in naturalistic contexts.

Such a benchmark has not yet been fully realized.

# 3. Experimental Methods

## 3.1. The W&H game and its adaptation for the WnH benchmark

The Watson & Holmes (W&H) tabletop game [5] was adapted for use in assessing the reasoning abilities of LLMs. We first describe the original form of the game, and then its adaptation.

Watson & Holmes is a whodunnit game, in which players compete to be the first to solve a mystery using reasoning skills such as deduction, induction, and abduction. Each play of the game involves a fresh case, which can only be attempted once only. Each case consists of:

- A case introduction given in narrative form (e.g. a report of a new client visiting Holmes). Mean length 600 words.
- Location cards with narrative text within which is information and clues to solving the case. Mean of 15 locations per case: mean of 160 words per location.
- 2-4 questions per case which players must answer correctly to solve it.
- Model answers to those questions.
- A case solution given in narrative form (e.g. Holmes describes his reasoning to Watson). Mean length 600 words.

As W&H is a commercial product that depends on players encountering each case for the first time, the case texts are neither distributed online nor provided in digital form with the game. We therefore consider it unlikely that they have appeared in any model training datasets.

**The unadapated game**

The unadapted game is designed for three to seven players. It begins with reading of a case introduction (e.g. Figure 1), review of the list of available case locations (e.g. Figure 2), and distribution of initial resources to the players, namely Carriage Tokens, which can be used to bid for locations. Play then proceeds with alternating visiting and investigation phases.

During visiting phases, each player in turn chooses a location to visit and places a bid on it using Carriage Tokens. If a location is already taken, any other player wishing to enter during that turn must outbid the current occupant. Once a visiting phase is finished, the game enters an investigation phase, where players privately review the text for the location they have chosen to visit, taking notes as desired, and follow any special conditions or restrictions indicated for the location.

When a player feels confident that they understand the mystery they can visit a special location (221B Baker St) to unlock the case questions (e.g. Figure 3, left). They write answers to the questions which they then compare to the model answers (e.g. Figure 3, right). If their answers are correct, they win the game; otherwise, they reveal only how many questions they answered correctly, and play continues for any remaining players.

Once a player has won the game, or all players have lost, players finish by reading the case solution (e.g. Figure 4) which presents an idealized reasoning path in narrative form.

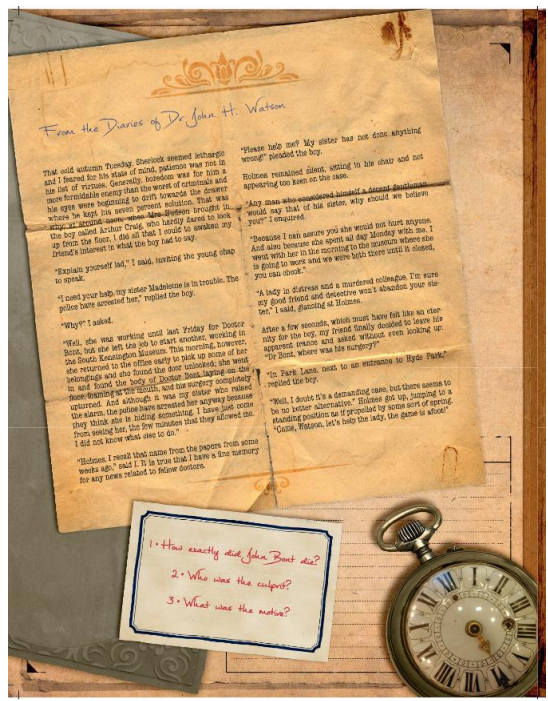

Figure 1: The introduction to the case 'A damsel in distress' as it is presented to the players in the tabletop game. Note, in this tutorial case (available free online) players get the questions along with the introduction, rather than only once they declare themselves ready to solve the case by visiting a special location.

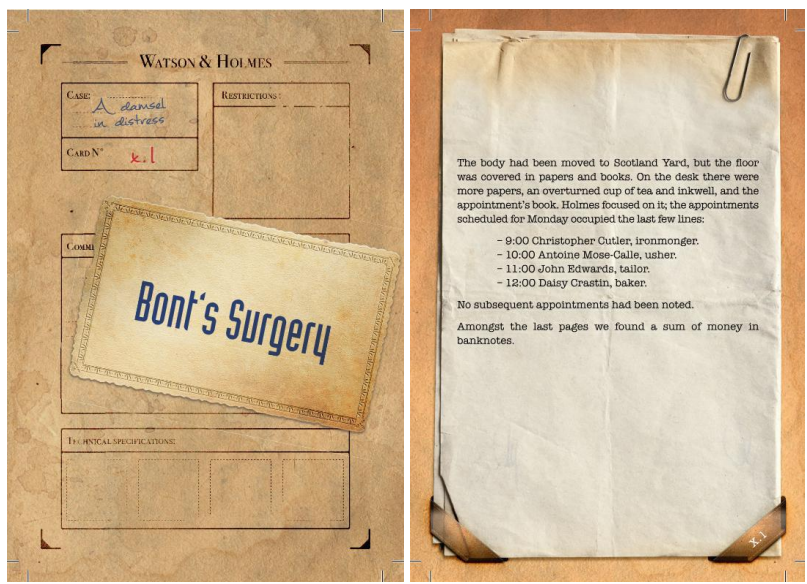

Figure 2: The location card 'Bont's Surgery` in the case 'A damsel in distress'. Left: card front, visible when choosing location to visit. Right: card back, viewed only when a player visits that location.

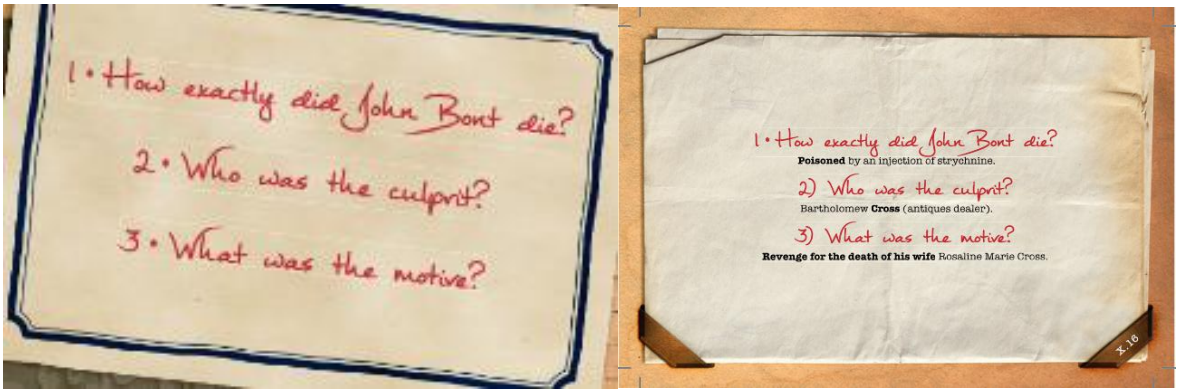

Figure 3: The questions (left) and model answers (right) to the case 'A damsel in distress'.

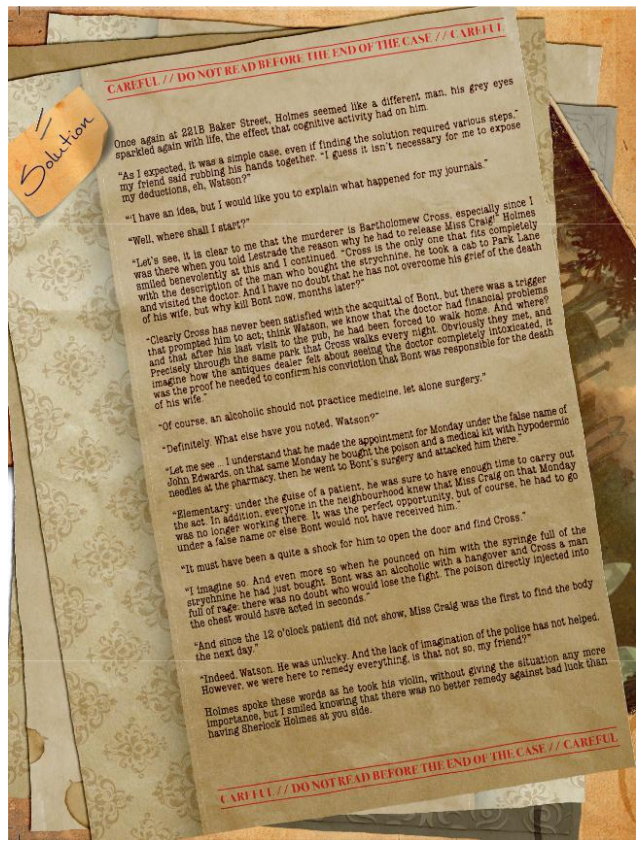

Figure 4: The case solution to the case 'A damsel in distress' as it is presented to the players in the tabletop game.

**The adapted game**

Beyond the detective skills required to solve a case, the game requires some strategic decision-making: choosing how to bid carriage tokens to secure a visit to a location, and choosing when to risk revealing and attempting the case questions. These strategic decisions are more complex in some cases that have special rules (e.g. the possibility to lock a location making it difficult for other players to access). To adapt the game to be a test of LLM reasoning we remove the strategic elements by making the following changes:

- Single player
- Case questions detailed as part of case introduction.
- Locations can be visited in any order without bidding or locking mechanisms.
- Case questions are answered after the introduction and after each location.
- Play continues until all locations have been visited.
- Cases with significant special rules or visual clues are not used for LLM play – leaving 12 suitable cases. The first case is used for human player practice and is excluded from analysis.

The adapted process flow is shown in Figure 5. The same workflow is used for both human participants and LLMs.

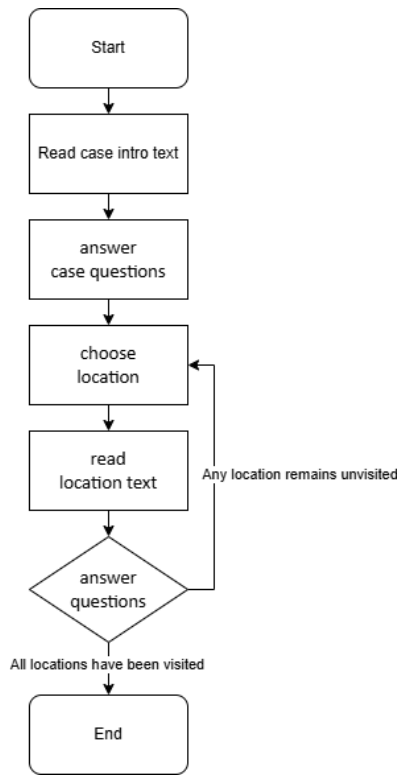

Figure 5: Adapted gameplay flow used for reasoning evaluation

The eleven cases used for performance evaluation had a mean of 15 (SD = 2) locations. The mean word count per case (introduction plus location texts) was 2,900 (SD = 600). The number of questions per case ranged from two to five (mean = 3.4), giving in total 37 questions across the eleven cases.

### 3.1.1. Human experiment

Five participants were recruited from UCL undergraduate Computer Science students subject to the criteria that they were at least 18 years old and had good English language proficiency. Solving detective cases was mentioned in the advertising to encourage motivated subjects.

Each participant solved each case in a separate online session which used video chat between the experimenter and participant, and an online slide deck and answer sheet. Participants were given access to the slide deck, which initially contained only the case introduction, the case questions, and a list of locations available for investigation. As play progressed the experimenter appended slides to the deck providing the narrative text of newly visited locations. Participants were able to navigate freely through the deck, revisiting previously viewed locations at their own pace. Participants were also free to take personal notes or highlight content within the slide deck as they progressed through the case. Players were given access to the answer sheet into which they recorded their answers to the case questions after reading the case introduction and after visiting each location. At each step, participants were encouraged to integrate newly revealed clues with those previously collected before submitting their next response.

The first case (case 0) was used to familiarize participants with the game and our implementation of it. The scores from this introductory case were not used in computing player performance.

Participants were rewarded according to a two-part scheme which incentivized diligent performance. First, participants were compensated at the London Living Wage rate (2025) for *time spent* solving cases (hence no incentive to rush). Second, the *highest-scoring* participant on each case received double payment for that case.

The UCL Computer Science Research Ethics Committee approved the experiment.

### 3.1.2. AI experiment

To implement the game for AI models we use three prompt templates as outlined below and detailed in appendix B.1. Each template is filled in with case- or location-specific information depending on the stage of play.

**Introduction prompt**. This prompt instructed the AI to solve the detective case and presented the case introduction and the case questions. The prompt ends with an instruction to the model to provide initial answers, in JSON format, to the questions.

**Choose location prompt**. This prompt recaps the case introduction (including the case questions) and the text of each location that has been visited so far during solving. This is followed by the list of locations not yet visited from which the model is instructed to select one and output it in JSON format.

Note that recapping the introduction and the text of locations already visited is necessary as the model is freshly prompted each time, rather than appending to a growing context. This prevents the model's previous answers to the case questions being visible to it as it makes a location choice, or answers case questions.

**Question answering prompt.** This prompt recaps the case introduction (including the case questions) and the text of each location that has been visited so far during solving. The prompt ends with an instruction to the model to provide answers, in JSON format, to the questions.

The models employed in the study are shown in Table 1, which includes both completion model (cModel) and reasoning (rModel) systems. Variants designed to reflect different levels of reasoning effort were also included (i.e., GPT-5-high vs. GPT-5, and o3-high vs. o3-pro). A default temperature value of 1.0 was applied when available, except for models that do not support temperature control (e.g., o1), for which the feature was disabled.

| LLM Model Name | Organization | Release Date | Model Type |
|---|---|---|---|
| Claude 3.5 Haiku | Anthropic | 22-10-2024 | cModel |
| Claude 3.7 Sonnet | Anthropic | 24-02-2025 | rModel |
| Claude 4 Opus | Anthropic | 22-05-2025 | rModel |
| Claude 4 Sonnet | Anthropic | 22-05-2025 | rModel |
| DeepSeek V3 | DeepSeek | 26-10-2024 | rModel |
| GPT-4.1 | OpenAI | 14-05-2025 | cModel |
| GPT-4.1 Nano | OpenAI | 14-05-2025 | cModel |
| GPT-4.1 Mini | OpenAI | 14-05-2025 | cModel |
| GPT-4o | OpenAI | 13-05-2024 | cModel |
| GPT-4o Mini | OpenAI | 18-07-2024 | cModel |
| GPT-5 | OpenAI | 07-08-2025 | rModel |
| GPT-5 Mini | OpenAI | 07-08-2025 | rModel |
| GPT-5 Nano | OpenAI | 07-08-2025 | rModel |
| GPT-o1 | OpenAI | 12-09-2024 | rModel |
| GPT-o3 | OpenAI | 16-04-2025 | rModel |
| GPT-o3 Mini | OpenAI | 16-04-2025 | rModel |
| GPT-o3 Pro | OpenAI | 10-06-2025 | rModel |
| Gemini 2.5 Flash | Google | 17-06-2025 | rModel |
| Gemini 2.5 Pro | Google | 17-06-2025 | rModel |
| Llama 4 Scout | Meta | 05-04-2025 | cModel |
| Llama 4 Maverick | Meta | 05-04-2025 | cModel |
| Mistral 7B Instruct v0.3 | Mistral | 27-09-2023 | cModel |

Table 1: Overview of LLM Models with Release Dates and Model Categories

## 3.2. Evaluating performance

### 3.2.1. Scoring System

In the unadapted version of the game, players answer the case questions only once, and the game is then ended for that player. In the adapted game, the player's answers are collected after the case introduction and after each location is visited.

In the unadapted game, answer correctness is determined by the player themselves, by comparison with the model answers provided, and binary scoring of each answer as correct or incorrect. In the adapted version each answer was scored on a 4-point scale. This allows intermediate levels of solving to be gauged, yielding a more precise score. The scoring scale was as follows.

> 0: No elements are present and correct, or the answer is misleading, nonsensical, or intentionally wrong.

> 1: Some elements are present and correct, but most are absent or incorrect.

> 2: Most elements are present and correct, but some are missing or incorrect.
>
> 3: All elements are present and correct.

In the original game all that matters is whether or not a player answers all questions correctly at their single opportunity to answer them. For the adapted game four score metrics are used.

- *Instantaneous* score is the score at a particular number of locations into a case.
- *Progressive* score is the instantaneous score averaged across all stages of solving a case.
- *Final* score is instantaneous score at the end of the case.
- *Overall* score is the mean of progressive and final scores. By combining progressive and final scores, a high overall score requires a case to be fully solved, and as early as possible.

Finally, *overall performance* was defined as the mean of the overall score across all questions in all cases.

### 3.2.2 Auto grading

As manual marking of case answers is time-consuming and potentially affected by the grader and their grading history, an automated AI-based grading system was developed to avoid these issues. To validate the autograder, a test dataset of sample answers was written by the experimenters, and these were graded by five markers working independently. For each question, these markers were provided with 2-3 example answers with grades at each level of grading 0-3. These grading examples were devised by the experimenters following discussion.

We used GPT 4.1 as autograder and experimented with the following variant prompting schemes (examples are provided in the appendix A.1.). All prompt modes included the case introduction, the case solution, the question, the model answer, and the task instruction. The differences between the modes were as follows:

- Mode 1: standard elements plus marking scheme and marking examples.
- Mode 2: same as mode 1, but elements in a different order.
- Mode 3: as mode 1 but without case solution
- Mode 4: standard elements plus an LLM-generated marking rubric.
- Mode 5: as mode 4 but without use of the case solution during rubric generation.

The rubrics in Modes 4 and 5 were generated by prompting GPT-4.1 to provide detailed guidance on how to assign scores. The inputs to this rubric generation process were the question, the model answer, marking examples and, for mode 4, the case solution. The generated rubric defined criteria for each score (0–3), including a definition of how the score should be awarded, examples of answers at that level, and edge-case guidance such as how to handle partial answers or minor misspellings.

To identify the best-performing autograding mode the grades assigned by each were compared to those assigned by the human graders, with the agreement quantified by the percentage of grades agreeing exactly (% Agreement), and the root mean square (RMS) difference in grades.

Table 2 details how the different autograder modes agree with human graders, giving the mean human-human agreement scores as comparison. Taking account of both agreement percentage and RMS grading mismatch, mode 5 was the best performing. Mode 5 autograder-human agreement is close enough to human-human agreement to make its use acceptable, and so all performance results in the remainder of this report will be based on mode 5 autograding performed by GPT 4.1.

| Comparison | % Agree | RMS |
|---|---|---|
| Human-Human | 88.6 | 0.40 |
| Mode 1-Human | 86.5 | **0.40** |
| Mode 2-Human | 85.2 | 0.44 |
| Mode 3-Human | 85.5 | 0.43 |
| Mode 4-Human | 86.8 | 0.41 |
| Mode 5-Human | **87.4** | 0.41 |

Table 2: Agreement metrics between autograder modes and human graders. % Agreement is the proportion of questions on which they produced the exact same answer grading, and RMS the root-mean-square error between their answer grades.

## 3.3. Supplementary Methods

### 3.3.1 Question IDA rating

For use in performance analysis, each question was rated according to the importance of the three types of reasoning – induction, deduction and abduction (IDA) – needed to answer it. Ratings were very low, low, medium, high, very high. These ratings were set by an AI (GPT-5 Thinking) based on the full text of a case. The prompt used was:

> "I am going to upload a document describing a Sherlock Holmes detective case, including an introduction, clues at different locations, and a solution. Three questions and their answers are provided. For each question, discuss the reasoning required to answer it correctly. Then I would like you to grade the importance of the three types of reasoning - inductive, deductive and abductive - in correctly answering the question. Grade each as very low, low, medium, high, very high."

Examples of the reasoning behind IDA ratings are provided in Figures 6.

> **Question 1: How exactly did John Bont die?**
>
> **Reasoning discussion**
>
> Answering this question requires **integrating medical, forensic, and contextual clues** rather than relying on a single explicit statement.
>
> Key observations include:
>
> - **Foam around the mouth**, consistent with certain fast-acting poisons.
> - **A puncture wound on the chest**, aligned with a small red stain on the shirt.
> - **The impossibility of oral strychnine ingestion** without detection, due to its bitter taste.
> - **Recent purchase of strychnine and a medical kit**, implying injectable administration.
>
> The crucial step is recognizing that **poisoning occurred via injection**, not ingestion. This eliminates common poisoning routes (tea, food) and points to **hypodermic delivery**, which explains both the puncture wound and the rapid onset of symptoms.
>
> This conclusion is reached by:
>
> - **Abductively inferring** the most plausible cause of death that explains all symptoms.
> - **Deductively ruling out alternatives** (oral poisoning, accidental exposure).
> - **Inductively drawing on general medical knowledge** about strychnine's effects.
>
> **Importance of reasoning types (IDA)**
>
> - **Inductive:** *Medium*
> - **Deductive:** *High*
> - **Abductive:** *Very high*

Figure 6: An example of the reasoning behind IDA rating.

The same rating procedure was repeated but with only the case introduction provided, so without the locations' text, yielding a second set of IDA ratings.

The induction, deduction, and abduction ratings across the eleven cases are summarized in table 3.

|           | Ratings (full case) | Ratings (just intro) |
|-----------|---------------------|----------------------|
| Induction | 1.1 (0.6)           | 1.1 (0.5)            |
| Deduction | 2.4 (0.9)           | 1.0 (0.9)            |
| Abduction | 3.1 (0.7)           | 2.9 (0.9)            |

Table 3: IDA ratings of questions IDA ratings – mean(sd), all on a scale of 0 to 4.

### 3.3.2 Choice of main models and assessment of their performance variance

To simplify results presentation, a high performing cModel and a high performing rModel were chosen for extra analysis.

- Main cModel: GPT-4.1
- Main rModel: GPT-5

AI performance may vary across solving attempts due to stochastic variation in internal reasoning and output token generation. To assess this, five repeat complete runs were made with the main models. In the results and discussion, the average score across the five runs are reported for the main models, and single run scores for other models.

### 3.3.3 Uncertainty Analysis

To assess uncertainty arising from the limited sample of human participants (n = 5), the finite set of WnH questions (n = 37), and variability across repeated model executions (five independent runs for the main rModel and cModel), a non-parametric bootstrap analysis was performed. This approach allows us to quantify (i) model performance uncertainty, and (ii) confidence in the comparison between model and human performance. By bootstrapping different combinations of study dimensions (participants, repeat runs and question pool) we determined the main contributors to the uncertainties.

### 3.3.4 Choice of prompt system on the model performance

To assess the impact of prompt design on LLM performance, the following prompt systems were used:

- Simple Prompt, used in the main AI experiment section and other experiments in this article, directly asks the model to provide an answer either after the case introduction or after visiting a location.
- Discussion Prompt, which is similar to the Simple Prompt, but additionally encourages the model to discuss and reason about the case before providing an answer. Accordingly, both the introduction and question-answering prompts are modified.
- Revision Prompt, which extends the Discussion Prompt by adding a further prompt that instructs the model to reflect on and critique its answer after the initial discussion, and then produce a final answer.

The full prompts are provided in Appendix C.1.

# 4. Results

## 4.1. Benchmark Scores

Table 4 and Figure 7 present the performance scores for all humans and models tested.

Focusing on the overall performance score (average of progressive and final), the best performing cModel was GPT-4.1 with a score of 1.45, slightly ahead of the mean human score of 1.43. GPT-4.1 is our main cModel that we focus analysis on. The best performing rModel was o3-pro with a score of 2.14, which is well ahead of the best human player's score of 1.80. Our main rModel, GPT-5, performed slightly below o3-pro with a score of 2.10. Assuming that the population of human scores is normally distributed, we estimate that the o3-pro and GPT-5 overall performance scores are in the top 5% of human scores.

| Model name | Progressive Score | Final Score | Overall score | Estimated Human Percentile |
|---|---|---|---|---|
| Player1 | 1.29 | 2.30 | 1.80 | 88 |
| Player5 | 1.20 | 2.03 | 1.61 | 72 |
| Player4 | 1.05 | 2.05 | 1.55 | 65 |
| Player3 | 0.91 | 1.68 | 1.29 | 33 |
| Player2 | 0.66 | 1.11 | 0.89 | 4 |
| Average Human | 1.02 | 1.83 | 1.43 | 50 |
| o3-pro | 1.81 | 2.46 | 2.14 | 99 |
| GPT-5 (high) | 1.81 | 2.41 | 2.11 | 98 |
| GPT-5 (main rModel) | 1.79 | 2.41 | 2.10 | 98 |
| o3 | 1.72 | 2.35 | 2.04 | 97 |
| o3 (high) | 1.72 | 2.32 | 2.02 | 97 |
| Gemini 2.5 Pro | 1.47 | 2.14 | 1.80 | 88 |
| o1 | 1.35 | 1.97 | 1.66 | 77 |
| Claude Opus 4 | 1.32 | 2.00 | 1.66 | 77 |
| Claude 3.7 Sonnet | 1.15 | 1.81 | 1.48 | 57 |
| GPT-4.1 (main cModel) | 1.05 | 1.86 | 1.45 | 53 |
| Claude Sonnet 4 | 1.13 | 1.76 | 1.45 | 52 |
| Gemini 2.5 Flash | 1.11 | 1.68 | 1.39 | 46 |
| GPT-5 mini | 1.00 | 1.59 | 1.30 | 34 |
| DeepSeek-V3 | 0.99 | 1.35 | 1.17 | 21 |
| GPT-4o | 0.87 | 1.32 | 1.10 | 15 |
| Llama 4 Maverick | 0.76 | 1.22 | 0.99 | 8 |
| Claude 3.5 Haiku | 0.73 | 1.03 | 0.88 | 4 |

| GPT-5 nano | 0.72 | 1.03 | 0.87 | 4 |
| GPT-4.1 mini | 0.59 | 0.73 | 0.66 | 1 |
| o3-mini | 0.56 | 0.68 | 0.62 | 1 |
| Llama 4 Scout | 0.50 | 0.59 | 0.55 | 0 |
| GPT-4.1 nano | 0.44 | 0.57 | 0.50 | 0 |
| Mistral 7B Instruct v0.3 | 0.43 | 0.46 | 0.44 | 0 |
| GPT-4o mini | 0.43 | 0.46 | 0.44 | 0 |

Table 4: Comparison of scores across models and participants. Within each group rows are ordered from high to low overall score. Key rows are shaded for clarity: blue for Human, orange for rModel, and purple for cModel.

As shown in Figure 7, rModels outperform cModels. The differences in performance between the top five rModels are small, and clearly ahead of other rModels. o3-pro performs best, closely followed by GPT-5-high, GPT-5, o3, and o3-high. Among the cModels there is a single distinct best model—GPT-4.1. Comparing the same rModel under different reasoning efforts (i.e., GPT-5-high vs GPT-5 and o3-high vs o3-pro) shows that additional reasoning effort yields only small improvements.

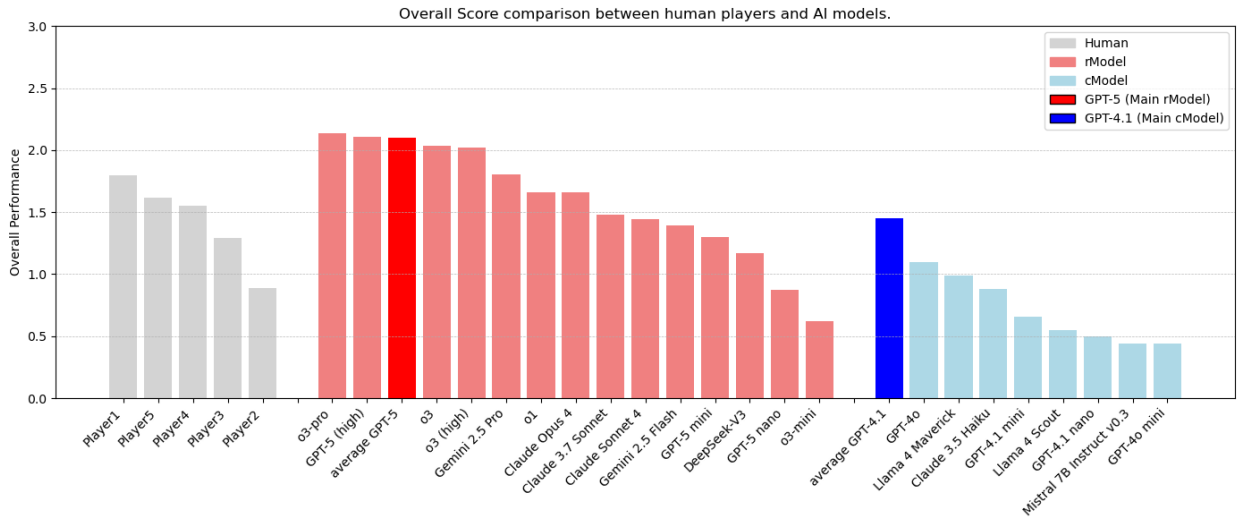

Figure 7: Visual comparison of overall scores for human players and AI models.

## 4.2. Uncertainty Analysis

**Model Performance**

The uncertainty of a single-run benchmark score arises from two sources: run-to-run variability and the finite (n=37) sample of WnH questions. Run variability was estimated from the five repeat runs of each of the main models to have a standard deviation of 0.08. Per-question overall score has a standard deviation of 0.89, yielding standard error of 0.15 given 37 questions. Combining these yields an overall one standard deviation of performance uncertainty of ±0.17 for performance at WnH-style questions when evaluated by a single run on our sample of 37 questions.

**Model vs Human**

The uncertainty of the performance of a model, when expressed as a percentile of the distribution of human scores, arises from run variability, finite question sample, and the finite (n=5) sample of humans tested. Table 5 shows 95% confidence intervals of the performance percentiles for the main models computed by bootstrapping these three dimensions of variability alone.

Considering the dimensions alone, and comparing the widths of the resulting confidence intervals indicates that the uncertainty due to the question and human samples are roughly equal and that due to run variability is smaller.

Considering the confidence intervals when all three sources of uncertainty are accounted for we see that while the best estimate of GPT-5 performance places it at the 98[th] percentile of human performance, a more cautious statement is that we are 95% confident that it is at the 86[th] percentile or higher. For GPT-4.1 our best estimate of performance places it at the 53[rd] percentile, but more cautiously we say that we are 95% confident it is at 91[st] percentile or lower.

| Bootstrap Dimensions | GPT-5 [95% CI] | GPT-4.1 [95% CI] |
|---|---|---|
| Human | 98% [93-100] | 53% [1-82] |
| Question | 98% [90-100] | 53% [34-71] |
| Runs | 98% [96-98] | 53% [45-61] |
| Human + Questions + Runs | 98% [86-100] | 53% [0-91] |

Table 5: Confidence intervals for model performance expressed as percentiles relative to the distribution of human scores.

## 4.3. Effect of Prompting Scheme

Table 6 indicates that varying the prompt system has no clear and strong impact on overall performance. There is a suggestion of a trend towards more complex prompting schemes improving performance, but given the uncertainty analysis result in section 4.5, which yielded that score differences below 0.17 are unlikely to be meaningful, this is not reliable.

Therefore, Simple Prompt is used as the default in all subsequent sections.

|  | GPT-4.1 | GPT-5 | average |
|---|---|---|---|
| Simple Prompt | 1.45 | 2.10 | 1.78 |
| Discussion Prompt | 1.60 | 2.02 | 1.81 |

| Revision Prompt | 1.50 | 2.18 | 1.84 |

Table 6: Effect of Prompt System Choice on Model Performance

## 4.4. Models vs. Humans: progressive solving

Fig 8 plots how instantaneous scores increase during solving cases. As we would expect these are generally increasing, as questions are answered better when more case narrative is available. For the main cModel (blue curves), the figure shows that the progression of instantaneous score is very similar to the average human (black curves). For the main rModel (red curves) the instantaneous score is higher than average human *even on the first round of question answering immediately after the case introduction*. The lead of the main rModel over the average human increases during the middle stages of the case, and slightly reduces at the final stages, but still finishes ahead of the average human.

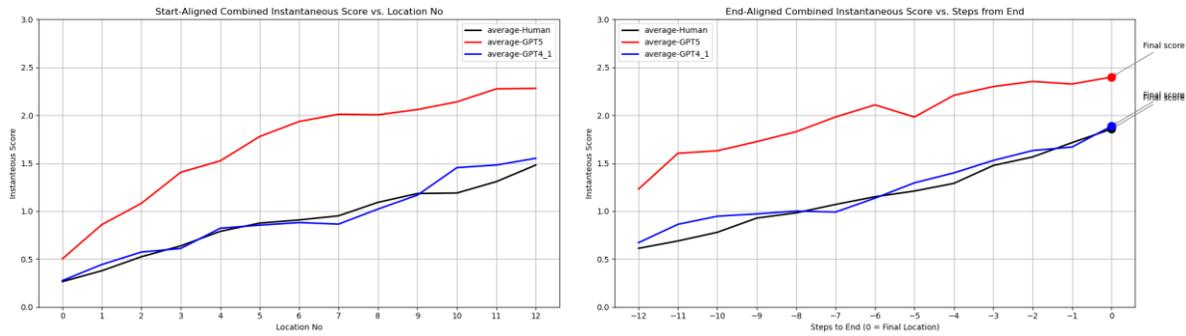

Figure 8: Start-aligned (left) and end-aligned (right) average instantaneous scores for Humans and the main models. Two plots are needed to accommodate the varying number of locations per case.

## 4.6. Models vs Humans: location ordering

Table 7 reports the correlations between the turn on which each location is visited in the solving of a case for two solvers. The results show that the main models are as similar to humans, as humans are to each other, in the order of visiting locations.

| Comparison | Correlation: mean (sd) |
|---|---|
| Human–Human | 0.44 (0.12) |
| GPT-5–Human | 0.52 (0.08) |
| GPT-4.1–Human | 0.48 (0.09) |

Table 7: Correlations describing the agreement in order of visiting locations between pairs of models and participants. Mean and sd are across all pairings of a type.

## 4.7. Models vs. Humans: question features

In this section we report the results for regressions on the *difference* in overall question score between the main models and the average human, which circumvents the confound of variable question difficulty.

Tables 8 present results for regressions of the difference in question overall score between main models and average human. The only significantly non-zero coefficients are the negative coefficients for word count for each of the main models vs. average human i.e. relative to humans, model performance declines on longer cases. Coefficients for IDA ratings are in no cases significantly non-zero.

| Feature | GPT-5 vs av. Human | GPT-4.1 vs. av. Human | GPT-5 vs. GPT-4.1 |
| --- | --- | --- | --- |
| word count | **-0.42 [-0.83, -0.03]** | **-0.43 [-0.78, -0.06]** | +0.01 [-0.41, +0.44] |
| I rating | +0.06 [-0.22, +0.34] | +0.01 [-0.26, +0.27] | +0.01 [-0.20, +0.29] |
| D rating | -0.02 [-0.29, +0.31] | -0.22 [-0.50, +0.06] | -0.22 [-0.50, +0.50] |
| A rating | +0.01 [-0.25, +0.32] | -0.05 [-0.31, +0.20] | +0.06 [-0.22, +0.38] |

Table 8: Regression coefficients for overall scores on questions

Additionally, we investigated whether there were any outlier questions where model overall scores deviated significantly from average human scores, even when adjusting for the systematic differences discovered by the regression analyses. Figures 9 depict the residuals between the model scores and the corresponding predictions from the regressions of model scores on human scores for GPT-5 vs. average Human and GPT-4.1 vs. average Human, respectively. For both regressions, the residuals were approximately normally distributed, and no significant outliers were identified.

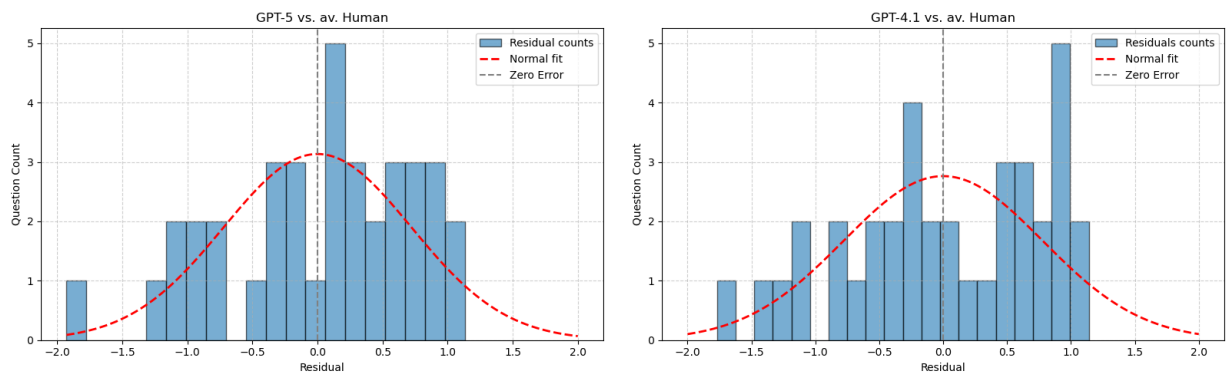

Fig 9: Histogram of Residuals. Left: GPT-5 vs. av. Human regression prediction. Right: GPT-4.1 vs. av Human – regression prediction.

Given the surprising advantage of rModels over human participants at the earliest stage of solving a case immediately after the case introduction (see section 4.7), we performed a separate regression analysis on the instantaneous score differences (main r-Model vs. human) at this stage. This regression used the IDA ratings for post-intro question answering, excluding a word count feature since the case introductions vary by only +/-10% in length.

As shown in Table 9, there is a positive coefficient for the I rating, statistically significant at the 90% level, but not for the D and A ratings. The interpretation is that the advantage of GPT-5 over average human, at the stage immediately post introduction, is greater when the Induction reasoning required for the question is greater.

| Feature | GPT-5 vs av. Human |
|---|---|
| I | **+0.28 [+0.08, +0.52]** |
| D | -0.09 [-0.29, +0.07] |
| A | +0.02 [-0.18, +0.26] |

Table 9: Regression coefficients for post introduction instantaneous scores on questions. 90% confidence intervals.

# 5. Discussion

## 5.1. Overall performance in context

Figure 10 visualizes the performance of the main models, with average and expert (95th percentile) human performance marked for context. The vertical scale of the figure charts overall score which is the average of final score and progressive score, where progressive score is itself the average of the instantaneous scores evaluated at each stage of solving a case. Consequently a maximum score of 3 is not achievable, even by a super-human reasoner, since most questions require information present on multiple location cards. We can compute a lower-bound on the maximum *achievable* score by averaging the per-question highest observed overall score, achieved by any participant or any model run. This yields a value of 2.45 which is marked on figure 11.

The key results of this study, visualized in figure 11, are:

- the main cModel (GPT-4.1) performed at average human level
- the main rModel (GPT-5) performed at expert (top 5%) human level.
- There is a sizable gap between the highest model score and the maximum achievable score.

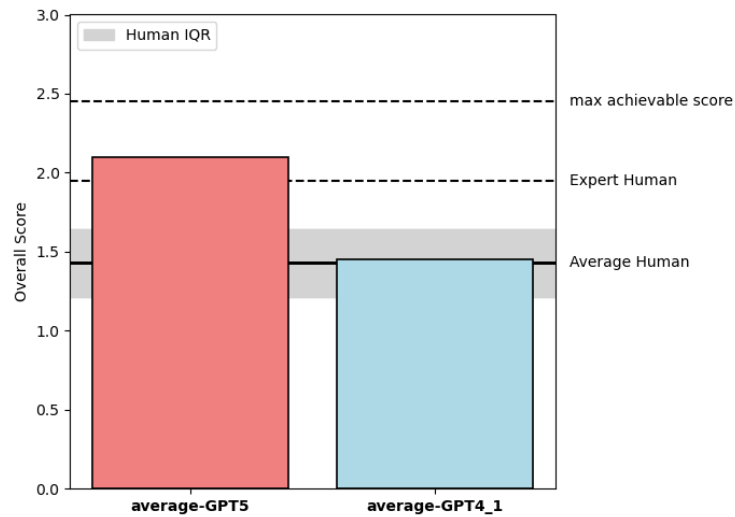

Figure 10: Performance summary of the main models and human performance.

Figure 11 puts the performance of the best models in temporal context, showing a trend of increasing performance with release date, and a step change with the extension from cModels to rModels.

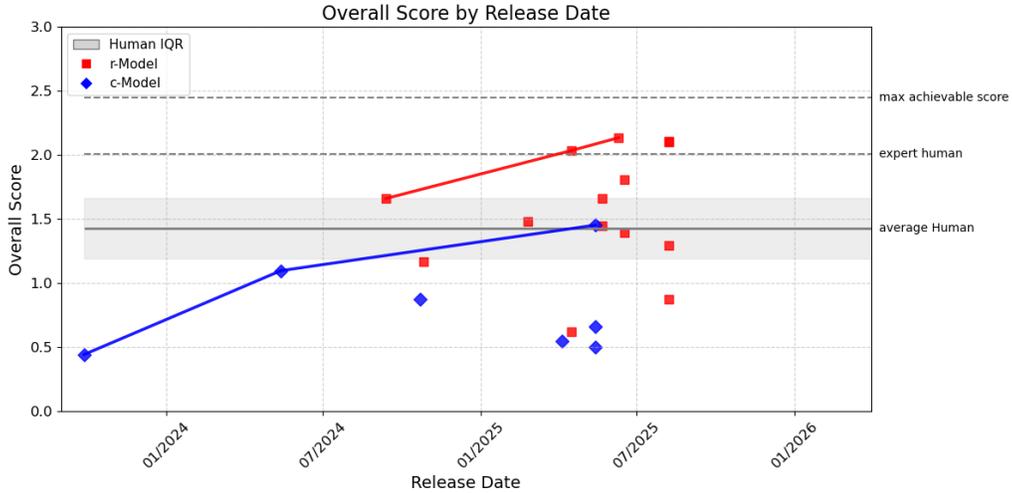

Figure 11: Overall scores of models by release date, compared against human performance. Lines added to indicate trends.

Figure 12 puts the performance of the best models in compute-cost context. The broad trend is of a linear increase in WnH performance as a function of $\log_{10}$ cost, but the deviations from the trend are of note. In particular the performance of GPT-5 is only slightly below that of the earlier o3-pro model from OpenAI, but is only 12.5% of the compute cost. Other models with relatively good performance given cost are: o3, Gemini 2.5 pro and Gemini 2.5 flash.

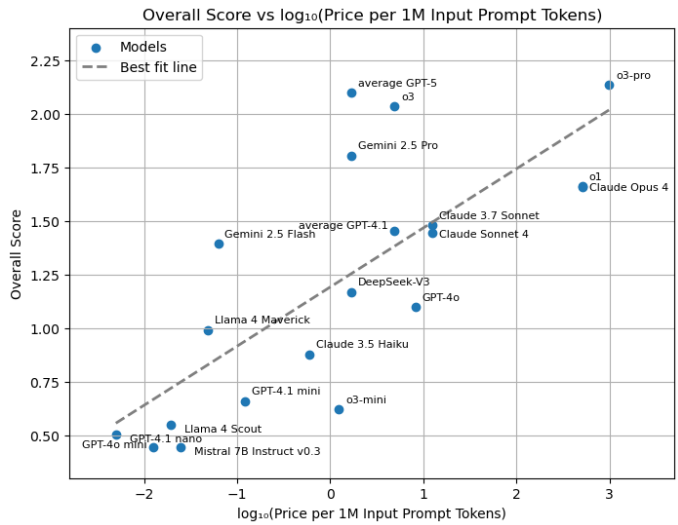

Figure 12: Performance against $\log_{10}$ (costs).

## 5.2. Quantitative Comparison with other benchmarks

Across detective-style benchmarks such as DetectBench, True Detective, and WHODUNIT the performance of cModels prior to GPT-4.1 is slightly behind human performance. While this appears consistent with the WnH benchmark, where GPT-4.1 performs at average human level, the lack of rigour in human testing of previous detective benchmarks limits the strength of conclusions that can be drawn.

To more clearly situate our benchmark within the broader evaluation landscape, we compare it against widely adopted more rigorous benchmarks: ARC-AGI-1 [77], ARC-AGI-2 [77], Humanity's Last Exam (Text Only) [62], and chatbot arena [96]. This comparison allows us to assess how performance on our benchmark relates to established measures of general reasoning and problem-solving ability.

Table 10 summarizes the pairwise correlations between benchmarks. This table is constructed by restricting the analysis to the set of models that are common across all chosen benchmarks. Overall, the correlations are moderate rather than strong, indicating only partial overlap in what each benchmark measures.

| Benchmark | WnH | hle | cba | ARC-AGI-1 | ARC-AGI-2 |
|---|---|---|---|---|---|
| WnH | 1.00 | 0.65 | 0.70 | **0.73** | 0.64 |
| hle | 0.65 | 1.00 | 0.55 | 0.78 | **0.84** |
| cba | **0.70** | 0.55 | 1.00 | 0.51 | 0.47 |
| ARC-AGI-1 | 0.73 | 0.78 | 0.51 | 1.00 | **0.87** |
| ARC-AGI-2 | 0.64 | 0.84 | 0.47 | **0.87** | 1.00 |

Table 10: score correlation between benchmarks. WnH denotes the proposed benchmark, HLE refers to Humanity's Last Exam, CBA denotes Chatbot Arena, and ARC-AGI-1 and ARC-AGI-2 refer to ARC-AGI versions 1 and 2, respectively. The strongest correlation in each row is bolded.

## 5.3. Discussion about the human participants

The online format proved effective for both participant engagement and reliable data collection.

The median time for human participants to solve a case was 59 minutes (IQR = [48, 71] mins). As shown in figure 13, Some players took systematically longer than others and this correlated with their overall performance ($R^2$=0.60, significantly non-zero at 95% confidence). The trend being that taking 50% longer to solve a case (from 50 minutes to 75 minutes) achieves, on average, an overall score 0.60 higher. This is a substantial effect given that the average overall score was 1.43.

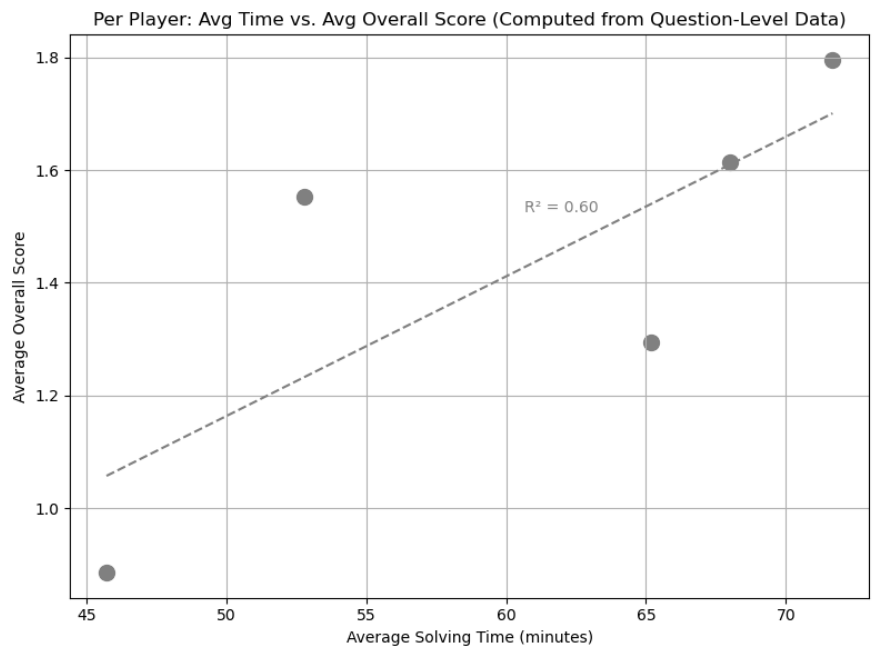

Figure 13: Plot showing the correlation between players' average case-solving time and their average overall score.

It is unclear whether longer time spent was a cause of better performance, or merely a correlate of greater natural aptitude. On reflection we note that the policy of awarding double payment to the winner in each case may have discouraged effort when a player felt they had little chance of winning. A more graded reward scheme might have encouraged more consistent effort.

Another consideration is whether more review and discussion of answers and grading after completing practice case 0 would have helped participants better understand how to improve their performance early in the solving of a case game, and how that early performance affected overall score.

## 5.4. Differences between model and human performance

Differences between model and human performance were observed linked to the word count of cases, the type of reasoning required, and the timecourse of solving cases.

### 5.4.1. Word Count

Longer cases were solved less well by models, relative to humans. This trend reached 95% significance for the main cModel and 90% significance for the main rModel. Figure 14 exhibits these trends in raw form. It is notable that these trends are apparent even though the range of cases lengths is 1900-4000 words, well within the claimed context lengths of these models. These findings suggest that claims of indefinitely expanding model contexts without performance degradation should be treated with caution: increasing length already appears to produce adverse impacts on LLM reasoning.

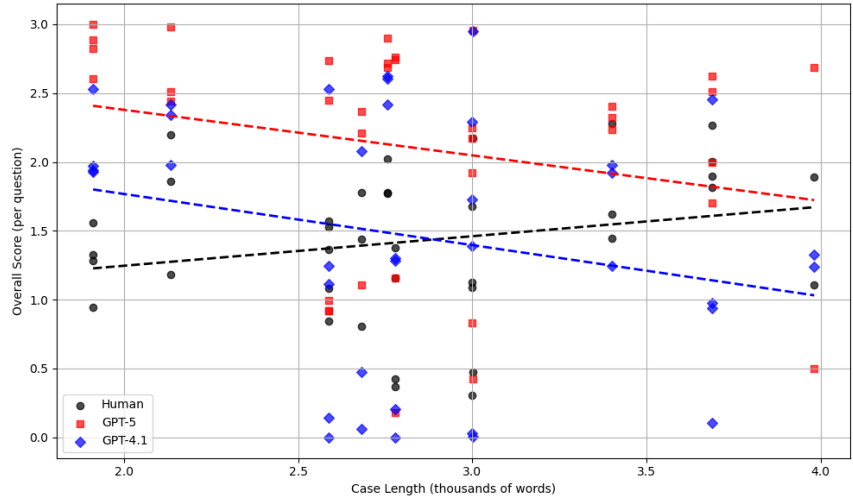

Figure 14: Overall Score by case length: Human vs rModel vs cModel. Lines are best fit trends.

### 5.4.2. Reasoning Type

For overall scores, reasoning type (IDA ratings) had no significant predictive power for the advantage of models over humans. However, for instantaneous scores, immediately after the introduction before any locations were visited, there was an advantage for GPT-5 on questions requiring more inductive reasoning (see table 11). Looking at the instances where this advantage manifested we see that the GPT-5 answers are bold leaps supported by multiple slight clues that are easily missed and integrated by human players.

It is worth noting that there are issues around the accuracy of the IDA ratings arising from the definitions of the reasoning modes themselves (section 2.1). While these definitions seem clear and distinct, rating their relevance for questions where all three modes are playing a role is very difficult, and the majority of questions in the naturalistic setting of the Watson & Holmes games are mixed mode. This ambiguity is not unique to our setting. Conan Doyle repeatedly presents Holmes's method as deductive, but numerous commentators have observed that many of Holmes's most celebrated "deductions" are in fact abductive inferences [97, 98].

## 5.4. Utility of the WnH benchmark

The WnH benchmark aims to be a broad assessment of model reasoning applied to natural tasks, precise enough to discriminate incremental improvements in reasoning capability. Our results give us some confidence that this has been demonstrated. We observe:

- Model performance ranging from below average to expert (top 5%) human level (fig 11).
- A trend in performance correlated with model release date (Figure 11).
- A step change in performance with the release of rModels (Figure 11).

Another strength of this benchmark is its pattern of correlations with other benchmarks. We visualize these using multi-dimensional scaling in Fig 15 which indicates that the WnH benchmark is distinct from other benchmarks, but not radically so.

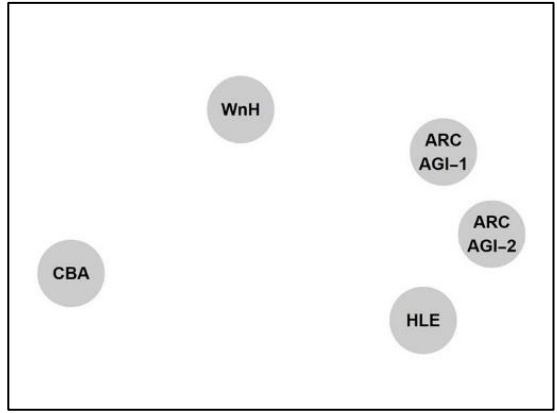

Fig 15 – MDS visualization of benchmark similarity. More correlated benchmarks are closer.

A weakness of our benchmark is that it is approaching the end of its utility for differentiating frontier models, as its performance saturates. Given score uncertainties of ±0.17 (section 4.2) the o3-pro score of 2.14 is significantly short of the maximum achievable score, which is at least 2.46 (see section 5.1). However, if model WnH performance continues its recent trend of improvement (figure 12) this gap will be closed by the end of 2026.

Benchmark saturation does not rob a benchmark of all utility. Current AI development is pursuing both high performance models designed for maximal reasoning performance [99], and small and efficient models optimized for deployability under tight resource constraints [100–102]. While a saturated benchmark is useless for tracking the performance of frontier models, it can remain useful for assessing deployable models.

# 6. Conclusion

This study developed and populated a novel reasoning benchmark (WnH) based on the Watson & Holmes tabletop detective game. Several key findings emerge.

First, AI expected performance at the benchmark has improved from lower quartile of our comparison human population (Computer Science undergraduates at a leading university) to top 14% level (95% confidence) over 9 months of 2025. Roughly half of that improvement is incremental with model release date, and half can be attributed to a step change when rModels were introduced.

Second, when model performance is considered in relation to compute cost, a clear efficiency trend emerges: newer models achieve higher reasoning ability at lower computational expense.

Third, rModels, compared to humans, are shown to have no clear strengths or weaknesses amongst the reasoning modes of inductive, deductive and abductive inference when the full case evidence is available; but there is an inductive reasoning advantage when incomplete evidence makes bold guesses based on multiple small hints the only route to answer. A clear result is that models reasoning performance decreases when evidence is presented across longer texts, even though these are considerably shorter than is claimed to be the working context length of contemporary models.

Finally, the utility of the benchmark is near its end for the frontier LLMs, with saturated performance expected to be achieved by the end of 2026. The benchmark is expected still to remain useful for assessing small, cost-effective AI models.

# 7. Appendix

## A.1. Autograding Prompts

### Mode 1

```
### Overview
You are an evaluator assessing the accuracy of answers to questions about a Sherlock Homes detective case. You will be given the case solution, a question and its model answer, and an answer to be assessed.

**The Case Solution:**
{case_solution}

**Question:**
{question}

**Model Answer:**
{model_answer}

### Marking Examples

Here are some examples of provided answers and the scores they received:

{marking_examples}

---

### Now assess the following answer:

**Provided Answer:**
{provided_answer}

### Your Task

Compare the provided answer to the model answer and to the case solution, and rate it on a scale of 0 to 3 according to the following criteria:
0: No elements are present and correct, or the answer is misleading, nonsensical, intentionally wrong
1: Some elements are present and correct, but most are absent or incorrect.
2: Most elements are present and correct, but some are missing or incorrect.
3: All elements are present and correct.

Make your assessment and return the result in **JSON format** as follows:
```json
{{
"SCORE": your_score_here
}}
```
```

## Mode 2

### Overview

You are an evaluator assessing the accuracy of answers to questions about a Sherlock Holmes detective case. You will be given the case solution, a question, its model answer, and an answer to be assessed.

### Definition of Levels 0-3

- **Level 0:** No elements are present and correct, or the answer is misleading, nonsensical, or intentionally wrong.

- **Level 1:** Some elements are present and correct, but most are absent or incorrect.

- **Level 2:** Most elements are present and correct, but some are missing or incorrect.

- **Level 3:** All elements are present and correct.

---

### Example Question and Model Answer

**Case Solution:**

{case_solution}

**Question:**

{question}

**Model Answer:**

{model_answer}

---

### Example Answers at Levels 0, 1, 2, 3

{marking_examples}

---

### Now assess the following answer:

**Provided Answer:**

{provided_answer}

### Your Task

Compare the provided answer to the model answer and to the case solution, and rate it on a scale of 0 to 3 according to the definitions above.

Return your assessment in **JSON format** as follows:

```json
{{
  "SCORE": your_score_here
}}
```

## Mode 3

### Overview

You are an evaluator assessing the accuracy of answers to questions about a Sherlock Homes detective case. You will be given a question and its model answer, and an answer to be assessed.

**Question:**

{question}

**Model Answer:**

{model_answer}

### Marking Examples

Here are some examples of provided answers and the scores they received:

{marking_examples}

---

### Now assess the following answer:

**Provided Answer:**

{provided_answer}

### Your Task

Compare the provided answer to the model answer and to the case solution, and rate it on a scale of 0 to 3 according to the following criteria:

0: No elements are present and correct, or the answer is misleading, nonsensical, intentionally wrong

1: Some elements are present and correct, but most are absent or incorrect.

2: Most elements are present and correct, but some are missing or incorrect.

3: All elements are present and correct.

Make your assessment and return the result in **JSON format** as follows:

```json
{{
"SCORE": your_score_here
}}
```

# Mode 4 – rubric writer

You are to generate a detailed, step-by-step marking rubric for assessing answers to Sherlock Holmes detective case questions.

Act as a world-class Sherlockian examiner: think aloud, compare clues, and lay out your criteria with detective-like reasoning.

You will receive the following inputs:

- Case Solution: A passage revealing the full answer and context of a Sherlock Holmes mystery, including the identity of culprits, methods, motives, etc.

- Model Answer: A concise ideal answer to the question, summarizing the key information from the case solution.

- Provided Answer: The answer to be assessed for accuracy and completeness.

- Marking Examples: Sample answers and the scores they should receive.

Your task:

- Generate a detailed JSON object describing the marking rubric for scoring the Provided Answer on a scale from 0 to 3.

For each score (0, 1, 2, 3), provide:

- A precise definition of when that score should be awarded, tailored to this case.

- At least 2–3 concrete examples (in the context of this case) of answers that would receive that score.

- Specific guidance on edge cases, including partial descriptions, aliases, and minor misspellings (which should NOT be penalized if the intended answer is clear).

Include a section titled "step_by_step_application" that explains, like a detective, how to apply the rubric in order to judge an answer. This should be in the style of a logical investigation: e.g., "First, compare the names... Next, look for essential details... Finally, consider if minor errors affect understanding."

Your rubric must:

- Be formatted as JSON with each score (0, 1, 2, 3) as a key, each mapping to an object with "definition", "examples", and "edge_case_guidance" keys.

- Include the "step_by_step_application" key with detective-style instructions for applying the rubric.

- Encourage careful, expert, thoughtful evaluation.

Emphasize: Minor misspellings should NOT be penalized if the intended answer is clear and unambiguous.

---

**The Case Solution:**

{case_solution}

**Question:**

{question}

**Model Answer:**

{model_answer}

**Marking Examples:**

{marking_examples}

---

**Return your rubric in the following JSON format:**

```json
{{
 "rubric": {{
  "0": {{
   "definition": "...",
   "examples": ["...", "..."],
   "edge_case_guidance": "..."
  }},
…..
  "3": {{
   "definition": "...",
   "examples": ["...", "..."],
   "edge_case_guidance": "..."
  }},
  "step_by_step_application": "..."
 }}
}}
```

# Mode 4, Mode 5 - rubric

### Overview

You are an evaluator assessing the accuracy of answers to Sherlock Holmes detective case questions. You will be given a question, a model answer, a detailed marking rubric, and the answer to be assessed.

**Question:**

{question}

**Model Answer:**

{model_answer}

**Marking Rubric:**

As a Sherlockian examiner, consult the rubric below. For each possible score, review the definitions, examples, and special guidance.

{rubric_json}

### Now assess the following answer:

**Provided Answer:**

{provided_answer}

### Your Task

Compare the provided answer to the model answer and the marking rubric above. Follow the rubric's step-by-step application instructions to assign a score from 0 to 3, considering the definitions, examples, and edge case guidance for each score.

Return your assessment in **JSON format** as follows:

```json
{{
  "SCORE": your_score_here
}}
```

# Mode 5 – rubric writer

You are to generate a detailed, step-by-step marking rubric for assessing answers to Sherlock Holmes detective case questions.

Act as a world-class Sherlockian examiner: think aloud, compare clues, and lay out your criteria with detective-like reasoning.

You will receive the following inputs:

- Model Answer: A concise ideal answer to the question, summarizing the key information from the case solution.

- Provided Answer: The answer to be assessed for accuracy and completeness.

- Marking Examples: Sample answers and the scores they should receive.

Your task:

- Generate a detailed JSON object describing the marking rubric for scoring the Provided Answer on a scale from 0 to 3.

For each score (0, 1, 2, 3), provide:

- A precise definition of when that score should be awarded, tailored to this case.

- At least 2–3 concrete examples (in the context of this case) of answers that would receive that score.

- Specific guidance on edge cases, including partial descriptions, aliases, and minor misspellings (which should NOT be penalized if the intended answer is clear).

Include a section titled "step_by_step_application" that explains, like a detective, how to apply the rubric in order to judge an answer. This should be in the style of a logical investigation: e.g., "First, compare the names... Next, look for essential details... Finally, consider if minor errors affect understanding."

Your rubric must:

- Be formatted as JSON with each score (0, 1, 2, 3) as a key, each mapping to an object with "definition", "examples", and "edge_case_guidance" keys.

- Include the "step_by_step_application" key with detective-style instructions for applying the rubric.

- Encourage careful, expert, thoughtful evaluation.

Emphasize: Minor misspellings should NOT be penalized if the intended answer is clear and unambiguous.

---

**Question:**

{question}

**Model Answer:**

{model_answer}

**Marking Examples:**

{marking_examples}

---

**Return your rubric in the following JSON format:**

```json
{{
 "rubric": {{
   "0": {{
     "definition": "...",
     "examples": ["...", "..."],
     "edge_case_guidance": "..."
   }},
   ….
   "3": {{
     "definition": "...",
     "examples": ["...", "..."],
     "edge_case_guidance": "..."
   }},
   "step_by_step_application": "..."
 }}}}```



## B.1. Full example AI experiment template

**Introduction Prompt**

OK you're going to try to solve a detective case.

When I give you instructions I will start the text with a ##. Text from the case will start with $$.

## Let's start with the case introduction.

$$ Introduction

[case_introduction_text]

## Based on the case information provided, give the most likely answer to each of the following [number] questions. If uncertain, make your best guess. Provide answers concisely, without explanations, in **JSON format**, where the **keys are string representations of the question numbers**. The format should look like this:

```json
{
  "1": "your_answer",
…
}
```

The questions are:

1. [question_text]

…



**Choose Location Prompt**

OK you're going to try to solve a detective case.

When I give you instructions I will start the text with a ##. Text from the case will start with $$.

## Let's start with the case introduction.

$$ Introduction

[case_introduction_text]

$$ Location: [location_name]:

[location details]

[visited_location_list]

## The questions to be solved are the following:

1. [question list]

…

## Choose the most informative location to visit next in order to solve the case efficiently. Select only **one**.

## Here are your options:

[Unvisited Location List]

Respond in JSON:

```json
{
  "LOCATION": your_location_here
}
```



**Question answering prompt**

OK you're going to try to solve a detective case.

When I give you instructions I will start the text with a ##. Text from the case will start with $$.

## Let's start with the case introduction.

$$ Introduction

[case_introduction_text]

$$ Location: [location_name]:

[location details]

[visited_location_list]

## Based on the case information provided, give the most likely answer to each of the following three questions. If uncertain, make your best guess. Provide answers concisely, without explanations, in **JSON format**, where the **keys are string representations of the question numbers**. The format should look like this:

```json
{
  "1": "your_answer",
…
}
```

The questions are:

1. [question_text]

…



## C.1. Full example of Discussion Prompt and Revision Prompt

**Discussion Prompt – Introduction**

OK you're going to try to solve a detective case.

When I give you instructions I will start the text with a ##. Text from the case will start with $$.

## Let's start with the case introduction.

$$ Introduction

[case intro text]

## Based on the case information provided, give the most likely answer to each of the following four questions. If uncertain, make your best guess. Provide answers concisely, without explanations, in **JSON format**, where the **keys are string representations of the question numbers**. The format should look like this:

```json
{
  "1": "your_answer",
  …
}
```

The questions are:

1. q1?

…

## You should discuss your analysis of the case before your answer. If relevant, discuss means, motive, and opportunity, and any unexplained aspect of the case evidence.



**Revision Prompt – Introduction Reflection**

OK you're going to try to solve a detective case.

When I give you instructions I will start the text with a ##. Text from the case will start with $$.

## Let's start with the case introduction.

$$ Introduction

[case intro text]

## Based on the case information provided, give the most likely answer to each of the following four questions. If uncertain, make your best guess. Provide answers concisely, without explanations, in **JSON format**, where the **keys are string representations of the question numbers**. The format should look like this:

```json
{
  "1": "your_answer",
  …
}
```

The questions are:

1. q1?

…

## You should discuss your analysis of the case before your answer. If relevant, discuss means, motive, and opportunity, and any unexplained aspect of the case evidence.

## Previous solver attempts reasoning and answer were as follows:

[previous answer]

## Please discuss and critique previous answer and please give your own answer which may be the same or different in JSON format.